\documentclass{article}

\usepackage[preprint]{neurips_2025}

\usepackage[hidelinks]{hyperref}
\usepackage{url}
\usepackage{booktabs}
\usepackage{amsfonts}
\usepackage{graphicx}
\usepackage{sidecap}
\usepackage[small]{caption}
\usepackage{subcaption}
\usepackage{amsmath}
\allowdisplaybreaks
\usepackage{amsthm}

\usepackage{wrapfig}
\usepackage{booktabs}
\usepackage{multicol}
\usepackage{float}
\usepackage{xcolor}
\usepackage{colortbl}
\usepackage{amssymb}
\usepackage{adjustbox}
\usepackage{pifont}
\usepackage{enumitem}
\usepackage[noabbrev,capitalize]{cleveref}

\usepackage{natbib}

\usepackage{appendix}

\usepackage[textwidth=2.7cm]{todonotes}

\newcommand\myshade{85}
\colorlet{myurlcolor}{blue}
\hypersetup{
  citecolor  = black,
  urlcolor   = myurlcolor!\myshade!black,
  colorlinks = true,
}

\title{DORAEMON: A Unifie\underline{D} Library f\underline{OR} Visu\underline{A}l Obj\underline{E}ct \underline{M}odeling and Representati\underline{ON} Learning at Scale}

\author{
  Ke Du, Yimin Peng, Chao Gao, Fan Zhou, Siqiao Xue\thanks{Corresponding author.} \\
  Independent
}

\begin{document}

\maketitle

\begin{abstract}
  DORAEMON is an open-source PyTorch library that unifies visual object modeling and representation learning across diverse scales. A single YAML-driven workflow covers classification, retrieval and metric learning; more than 1 000 pretrained backbones are exposed through a \texttt{timm}-compatible interface~\citep{rw2019timm}, together with modular losses, augmentations and distributed-training utilities. Reproducible recipes match or exceed reference results on ImageNet-1K~\citep{imagenet15russakovsky}, MS-Celeb-1M~\citep{jin2018community} and Stanford online products~\citep{songCVPR16}, while one-command export to ONNX or HuggingFace bridges research and deployment. By consolidating datasets, models, and training techniques into one platform, DORAEMON offers a scalable foundation for rapid experimentation in visual recognition and representation learning, enabling efficient transfer of research advances to real‑world applications. The repository is available at {\small\url{https://github.com/wuji3/DORAEMON}}. 
\end{abstract}

\section{Introduction}
\label{section:intro}

Large-scale visual object modeling seeks to learn robust, universal representations from massive and heterogeneous image datasets. A unified backbone architecture enhances a model’s generalization and multi‑task adaptability across classification~\citep{imagenet15russakovsky,simonyan2014very,he2016deep}, detection~\citep{chu2024soradetector}, retrieval~\citep{garcia2018asymmetric,wang2023image}, and visual reasoning tasks~\citep{kolesnikov2020big,xue2024famma}.

\begin{figure}[H]
    \centering

    \begin{minipage}[t]{0.47\linewidth}
        \centering
        \includegraphics[width=\linewidth]{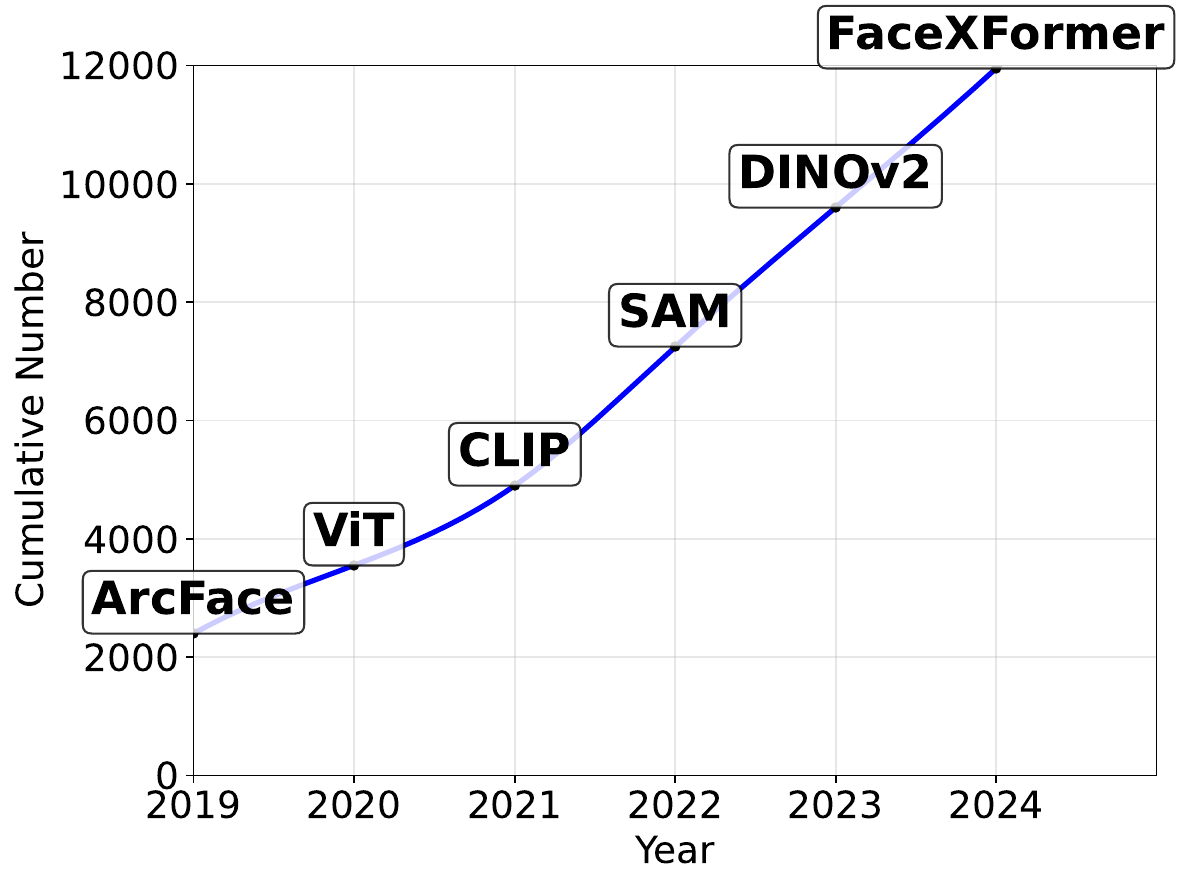}
        \vspace{2pt}
        \subcaption{Cumulative number of arXiv publications on large-scale visual object modeling. Labelled points highlight representative papers.}
    \end{minipage}
    \hfill
    \begin{minipage}[t]{0.47\linewidth}
        \centering
        \includegraphics[width=\linewidth]{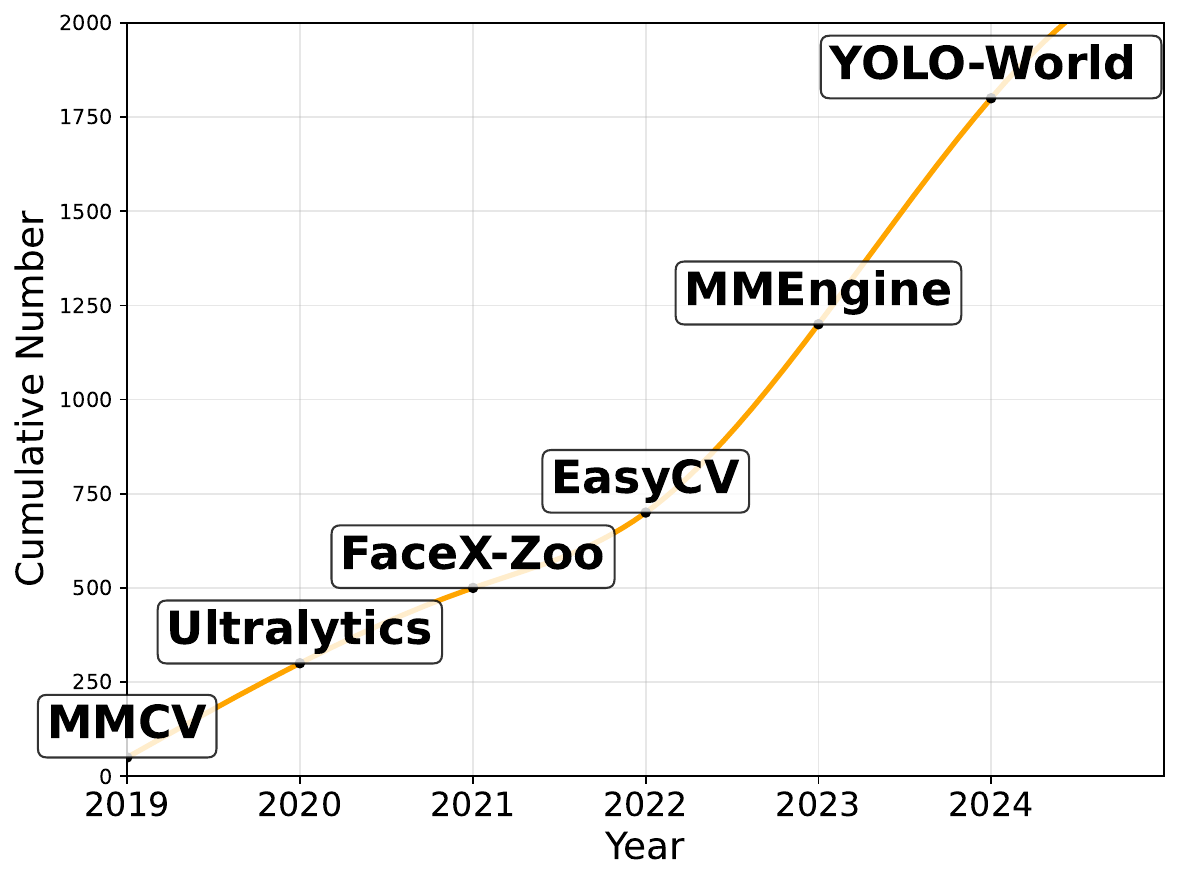}
        \vspace{2pt}
        \subcaption{Cumulative number of Github open-source projects for large-scale visual object modeling. Labelled points highlight representative repositories.}
    \end{minipage}

    \caption{The trends of large-scale visual object modeling (2019 - 2024).}
    \label{fig:trends}
\end{figure}

The rapid growth of research papers and publicly available codebases reflects the vitality of this field and the community’s increasing interest and progress, as illustrated in~\cref{fig:trends}. Meanwhile, large‑scale visual object modeling faces practical hurdles: fragmented codebases, task‑specific pipelines, and inconsistent training practices. Researchers often need to integrate disparate datasets, model architectures, and loss functions by hand, which complicates replication and slows experimentation. This lack of a unified framework makes it difficult to compare methods across tasks, underscoring the need for cohesive, scalable tooling.

To address these challenges, we have developed DORAEMON, a unified PyTorch framework aimed at supporting diverse tasks in large-scale visual object modeling, with the following key advantages:
\begin{itemize}[leftmargin=*]
    \item Comprehensive model support. Ready‑to‑use workflows for image classification, face recognition, and retrieval are paired with integration of over 1,000 pretrained architectures via \texttt{timm}, enabling efficient benchmarking across a broad range of backbones.
    
    \item Flexible training pipeline: Modern optimization algorithms with advanced regularization techniques---MixUp~\citep{zhang2018mixup}, focal loss~\citep{Detectron2018}, CutOut~\citep{devries2017cutout}, and more---together with dynamic data augmentation, layer‑specific learning rates, and hard‑example mining, provide fine‑grained control of the training process.

    \item Interpretability and deployment toolkit: Built‑in Grad‑CAM~\citep{Selvaraju_2019} visualizations facilitate qualitative analysis and debugging of learned representations without additional setup. Meanwhile, a seamless HuggingFace integration offer a clean interface for deploying models.
\end{itemize}

\section{Related Work}
\label{section:related_work}

The first generation of open-source libraries—most notably \texttt{timm}~\citep{rw2019timm}—standardised access to ImageNet-style backbones, yet left researchers writing bespoke code for data pipelines, training loops, and evaluation scripts. FaceX-Zoo~\citep{wang2021facex} later shipped a complete PyTorch stack for face recognition, but maintenance has slowed and the scope remains limited to a single task.
Recent projects such as EasyCV~\citep{easycv} and MMCV~\citep{mmcv} expanded coverage to self-supervised learning, detection, and segmentation. However, both repositories have seen only minor activity since 2023; as a result, recent algorithms and deployment utilities have yet to be updated.

DORAEMON brings a unified, YAML-driven framework that spans image classification, retrieval, and metric learning. It ships with (i) more than 1\,000 pretrained backbones; (ii) elastic, fault-tolerant distributed training; and (iii) one-click model export to TensorRT, ONNX, or TorchScript. By bundling task abstractions, dataset curation, and industrial-grade deployment into one coherent library, DORAEMON removes the residual friction between research prototypes and production-scale visual systems.


\section{Technical Background}
\label{section:background}

\paragraph{Definition.} Given an input image \( \mathbf{x} \in \mathbb{R}^{H \times W \times C} \), visual object modeling aims to extract compact, discriminative, and reusable representations, which can generalize across various downstream tasks. This process is formalized as visual representation learning, using a parameterized encoder \( f_\theta \) to extract a discriminative feature embedding \( \mathbf{z} = f_\theta(\mathbf{x}) \in \mathbb{R}^d \).

To support diverse tasks such as classification, recognition, and retrieval, modern visual systems typically decouple the shared representation encoder from task-specific output heads. Each task is equipped with a lightweight head \( h_\phi \), applied on \( \mathbf{z} \), and optimized with task-relevant losses. This modularity enables unified training, joint optimization, and scalable adaptation across tasks.

DORAEMON provides a unified framework \( \mathcal{F}_\theta \) that facilitates consistent and scalable representation learning across diverse visual tasks, including image classification~\citep{he2016deep,lecun1989backpropagation}, face recognition~\citep{schroff2015facenet}, and image retrieval~\citep{zhou2017recent}.

\vspace{0.5em}
\paragraph{Task Scenarios.}
Image classification aims to assign category labels to input images. It is typically solved via a linear classifier with cross-entropy loss~\citep{mao2023cross}, and is considered the cornerstone of supervised visual learning.

Face recognition seeks to verify or identify human identities. Unlike classification, it relies heavily on metric learning objectives such as ArcFace~\citep{deng2019arcface}, CircleLoss~\citep{sun2020circle}, and MagFace~\citep{meng2021magface} to enforce intra-class compactness and inter-class separation in the embedding space.

Content-based image retrieval (CBIR) retrieves semantically similar images based on visual content, aiming to distinguish between fine-grained visual instances. Instead of relying on category-level supervision, CBIR models are typically trained with metric learning objectives such as triplet~\citep{schroff2015facenet} or contrastive losses~\citep{radford2021learning,chen2020simple}, which encourage similar images to have closer embeddings and dissimilar ones to be farther apart.

\vspace{0.5em}
\paragraph{Unified Backbone.}
Despite differing objectives, classification, face recognition, and image retrieval share a common requirement: extracting rich and transferable visual representations. This observation motivates a unified modeling framework that reuses a shared encoder backbone across tasks. By leveraging a common representation space, models benefit from broader data coverage and improved generalization.

\vspace{0.5em}
\paragraph{Task-specific Heads.}
To adapt the shared backbone to diverse objectives, DORAEMON adopts a modular design with flexible prediction heads and task-specific loss functions. It integrates over 1,000 pretrained models via the \texttt{timm} ecosystem, supporting consistent representation learning across classification, face recognition, and image retrieval. This unified infrastructure enables seamless pretraining, fine-tuning, and evaluation—offering both conceptual elegance and engineering scalability.

Overall, the diverse objectives of classification, face recognition, and image retrieval are underpinned by a shared need for robust and transferable visual representations. By decoupling representation learning from task-specific prediction heads, modern visual systems achieve modularity and scalability. DORAEMON leverages this paradigm by providing a unified backbone architecture with flexible head configurations, enabling consistent and efficient learning across multiple vision tasks. This unified formulation not only streamlines model development but also lays the foundation for joint optimization and cross-task generalization, which are central to the design of our framework.

\section{Pipeline}
\label{section:pipeline}

\begin{figure}
    \centering
    \includegraphics[width=0.9\linewidth]{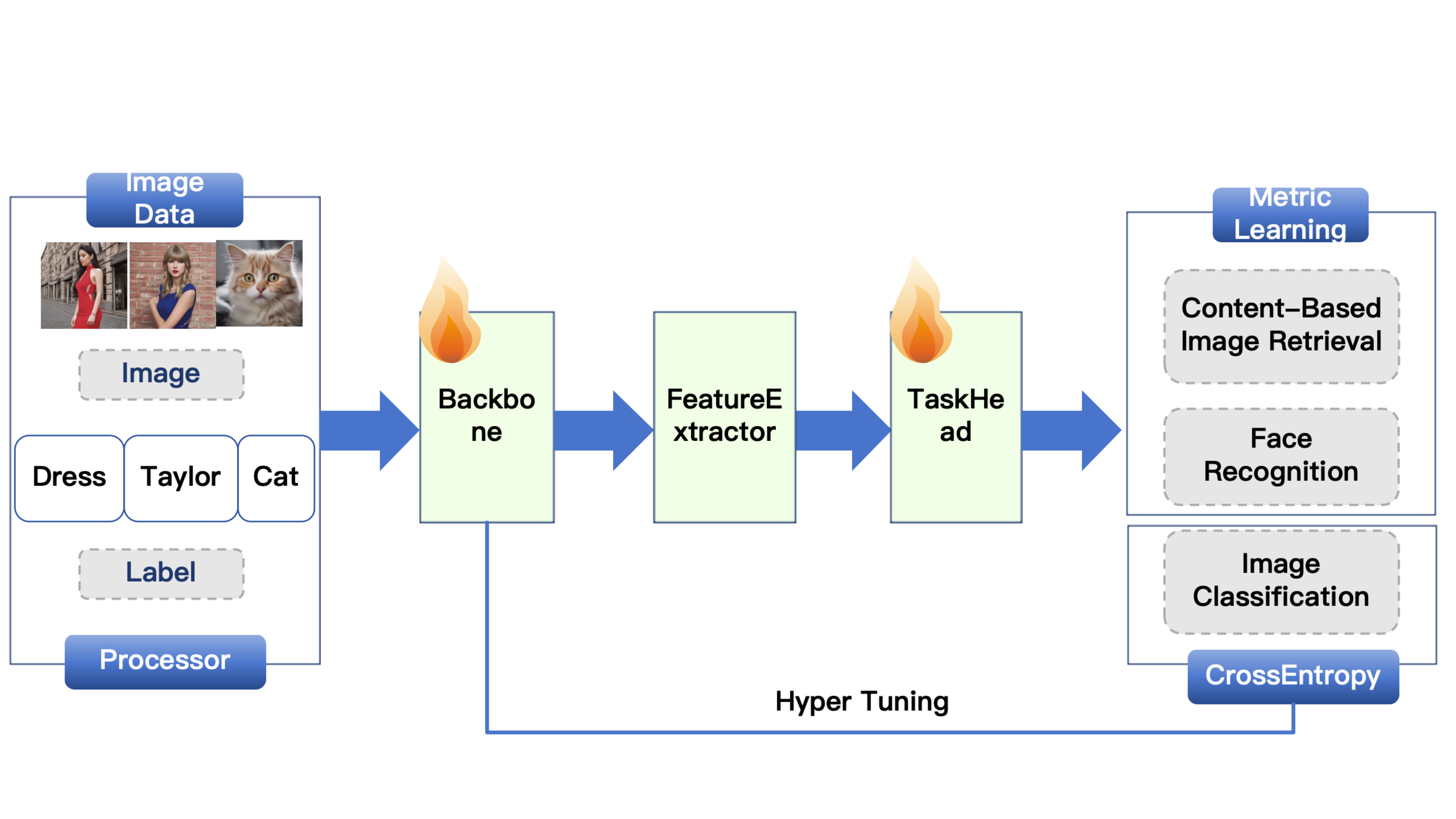}
    \caption{Unified training pipeline of DORAEMON. The framework uses a shared visual backbone with modular task heads for classification, face recognition, and retrieval. Data processing, loss functions, and optimization are all configurable via YAML for scalable deployment.}
    \label{fig:pipeline}
\end{figure}

Figure~\ref{fig:pipeline} presents the unified training pipeline of DORAEMON. The design centers around shared visual representation learning across tasks such as image classification, face recognition, and content-based image retrieval, by combining a common backbone with flexible task-specific heads. The system follows a modular flow consisting of data processing, representation extraction, task adaptation, and loss computation.

\paragraph{Data Preprocessing and Augmentation.}  
The input consists of raw images and their associated labels, which can be categorical (e.g., dress, cat) or instance-specific (e.g., Taylor Swift for face ID). Images are first passed through a task-agnostic processor that handles: Image decoding and resizing, Dynamic data augmentation pipelines defined in \texttt{utils/augment.py}, including RandomCrop, ColorJitter, CutOut~\citep{devries2017improved}, Copy-Paste~\citep{chen2020simple}, and class- or instance-aware augmentations. Label normalization and one-hot encoding for classification, or identity mapping for retrieval/recognition.
The augmentation intensity is scheduled via YAML flags such as \texttt{aug\_epoch} and \texttt{prog\_learn}, enabling curriculum-style regularization over training.

\paragraph{Unified Backbone.} 
All tasks share a common image encoder \( f_\theta \), implemented through the \texttt{timm} wrapper. Supported architectures include ResNet~\citep{he2016deep}, Swin Transformer~\citep{liu2021swin}, Vision Transformer~\citep{dosovitskiy2020image}, Masked AutoEncoder(MAE)~\citep{he2022masked}, and CLIP~\citep{radford2021learning}, totaling over 1,000 pre-trained variants. The encoder transforms the image \( \mathbf{x} \) into a semantic embedding \( \mathbf{z} = f_\theta(\mathbf{x}) \in \mathbb{R}^d \), which is further processed by downstream heads. Fine-tuning strategies allow freezing certain layers or fully adapting the encoder.

\paragraph{Task-specific Head.}
To support different downstream tasks, DORAEMON attaches a modular prediction head \( h_\phi \) on top of the shared representation \( \mathbf{z} \). For image classification, a standard Softmax classifier \( \hat{y} = \mathrm{Softmax}(h_\phi(\mathbf{z})) \) is used and optimized with cross-entropy loss~\citep{mao2023cross}. For face recognition, DORAEMON provides angular-margin heads such as ArcFace~\citep{deng2019arcface}, CircleLoss~\citep{sun2020circle}, and MagFace~\citep{meng2021magface}, which enforce angular separability in the embedding space. For content-based image retrieval (CBIR), triplet-based heads~\citep{schroff2015facenet} or cosine-margin modules~\citep{deng2019arcface} are adopted, optimized for instance-level similarity and Recall@K~\citep{radenovic2018revisiting} evaluation.

\paragraph{Loss Functions.}
The overall training objective consists of task-specific loss functions that are flexibly registered via YAML. For classification tasks, cross-entropy loss~\citep{mao2023cross} is defined as:
\[
\mathcal{L}_{\mathrm{CE}} = -\frac{1}{N} \sum_{i=1}^N y_i \log \hat{y}_i.
\]
For retrieval tasks, triplet loss encourages similar images to be embedded closer:
\[
\mathcal{L}_{\mathrm{triplet}} = \frac{1}{N} \sum_{i=1}^N \max\big(0, D(z_a, z_p) - D(z_a, z_n) + \alpha\big),
\]
where \( D(u,v) = \|u - v\|_2^2 \), and \( \alpha \) is a margin hyperparameter. For face recognition, angular margin losses (e.g., ArcFace) are used:
\[
\mathcal{L}_{\mathrm{Arc}} = -\frac{1}{N} \sum_{i=1}^N \log 
\frac{e^{s\cos(\theta_{y_i} + m)}}{e^{s\cos(\theta_{y_i} + m)} + \sum_{j \ne y_i} e^{s\cos(\theta_j)}},
\]
where \( \theta_j = \arccos(\mathbf{w}_j^\top \mathbf{z}_i) \), and \( s \), \( m \) are scale and margin.

\paragraph{Training Strategies and Configuration.}
Training is orchestrated through a modular hyperparameter system. 
DORAEMON supports optimizers including SGD~\citep{ruder2016overview}, Adam~\citep{adam2014method}, and Sharpness-Aware Minimization (SAM)~\citep{foret2020sharpness}, along with learning rate schedulers such as cosine annealing with warmup. 
Regularization techniques like label smoothing~\citep{szegedy2016rethinking}, focal loss~\citep{Detectron2018}, OHEM~\citep{shrivastava2016training}, and class-aware MixUp~\citep{zhang2018mixup} are integrated and controlled via YAML flags. 
Task-specific evaluation metrics (e.g., Top-1 accuracy, Recall@K, ROC AUC) are logged automatically during training to support efficient tuning.

\paragraph{Scalability and Integration.}
The pipeline is scalable to multiple tasks or datasets in a unified training session. DORAEMON supports both PyTorch checkpoint export and HuggingFace-compatible APIs:
\[
\texttt{AutoModel.from\_pretrained()},\quad \texttt{AutoProcessor.from\_pretrained()},
\]
ensuring seamless integration in academic research and real-world applications.

\section{Software Interface}
\label{section:interface}
The purpose of building DORAEMON is to provide a unified and extensible framework for large-scale visual object modeling and representation learning across multiple tasks such as classification, retrieval, and metric learning. 
For researchers, DORAEMON offers a modular interface to integrate and benchmark new backbones, losses, and training strategies with minimal effort, facilitating fair comparisons with existing methods. 
For industrial practitioners, DORAEMON includes a rich set of pre-configured pipelines and pretrained models, enabling rapid deployment and evaluation on real-world visual tasks. 
Full documentation and usage guides can be found in the repository.

\section{Ongoing and Future Work}
\label{sec:ongoing_future_work}

The rapid progress of large language models (LLMs)---especially their emerging ability to invoke external tools~\citep{xue2023weaverbird,chu2023leveraging,xue2024demonstration,zhou2024dbgpthub,huang2024romas}---opens new directions for our framework. We are pursuing two complementary strands:

\begin{itemize}[leftmargin=*]
\item Integration of powerful agents.  
Users may want our system not only to perform the training but also provide more powerful abilities, such as integration with database~\citep{xue2023dbgpt,xue2024demonstration}, support of time-stamped vision analytics and forecasting over historical image / video streams~~\citep{jin2023large,xue2022hypro} based on predictive decision abilities~\citep{pan2023deep}. These additions let users treat the library not just as a training toolkit but as an tool-using agent for vision workflows.

\item Advanced training paradigms for multimodal LLMs. Beyond the default vision-language pre-training pipeline, we will add built-in support for  
  (i) continual multimodal pre-training that incrementally ingests new text–image / video, and  
  (ii) lightweight prompt-based tuning across modalities~\citep{jiang2023anytime,xue2023prompttpp,jiang2025}.  
  These capabilities will let researchers prototype lifelong and task-specific multimodal LLMs with minimal overhead.

\end{itemize}

\section{Conclusion}

We have presented DORAEMON, a PyTorch-based library that delivers strong, ready-to-use baselines for image classification, face recognition, and content-based image retrieval.  Its modular APIs, extensive data–augmentation suite, and built-in visualization tools lower the entry barrier for computer-vision research while remaining highly extensible for large-scale production workloads.  

Looking ahead, the roadmap in Section~\ref{sec:ongoing_future_work} positions DORAEMON as a springboard for multimodal large-language-model research.  We hope the community will adopt and extend it to accelerate progress toward robust, decision-aware multimodal systems.

\bibliographystyle{icml2020_url}
\bibliography{reference}

\end{document}